\documentclass[english,12pt, letter]{article}

\usepackage{amssymb,amsmath,amsthm,graphicx} 
\usepackage[numbers,sort&compress]{natbib}
\usepackage{microtype}
\usepackage{subfigure}
\usepackage{comment,amsfonts,mathtools,bm,booktabs,lmodern, microtype,xcolor,algorithm}
\usepackage{enumerate}
\graphicspath{ {./} {./figures/} }
\usepackage[noend]{algorithmic}
\usepackage{hyperref}
\hypersetup{colorlinks=true,citecolor=red}
\usepackage{geometry}
\usepackage{autobreak}
\usepackage{bbm}
\providecommand{\keywords}[1]
{
  \small	
  \textbf{\textit{Keywords---}} #1
}

\begin{document}
\title{OWA aggregation of multi-criteria\\ with mixed uncertain fuzzy satisfactions}

\author{Yunjuan Wang \\ School of Computer Science \\ Xi'an Jiaotong University \\ email: \url{wangyunjuanxjtu@163.com} \and Yong Deng \\ Institute of Fundamental and Frontier Science \\ University of Electronic Science and Technology of China  \\ email: \url{dengentropy@uestc.edu.cn}; \url{prof.deng@hotmail.com} }

\date{June 2018}
\maketitle

\begin{abstract}
 We apply the Ordered Weighted Averaging (OWA) operator in multi-criteria decision-making. To satisfy different kinds of uncertainty, measure based dominance has been presented to gain the order of different criterion. However, this idea has not been applied in fuzzy system until now. In this paper, we focus on the situation where the linguistic satisfactions are fuzzy measures instead of the exact values. We review the concept of OWA operator and discuss the order mechanism of fuzzy number. Then we combine with measure-based dominance to give an overall score of each alternatives. An example is illustrated to show the whole procedure.

\keywords{ OWA operator; Multi-criteria; Triangular fuzzy numbers; Measure-based dominance; Decision-making}
\end{abstract}

\section{Introduction}
\label{Introduction}
How to make a proper decision to satisfy multi-criteria has attracted much attention over the last few years \cite{liu2018dynamic,ng2008efficient,karpak2001purchasing,narasimhan2006multiproduct,peng2014OWA,yager2017generalized}. Using OWA operator is always considered to be an efficient approach to aggregate the degree of satisfactory in individual criteria, thus offering a measure to order a collection of alternatives and choose the most satisfactory of it \cite{gul2018owa,yager2017owa,mesiar2018aggregation,yan2008fuzzy,liu2004preference,fonooni2007Rational,zhao2013Sensitivity,xu2011Intuitionistic,zarghami2008new,zarghami2008fuzzy}. OWA operator has been widely used in many applications such as decision analysis \cite{rinner2002Web,yusoff2015analysis,pelaez2007analysis}, fuzzy logical control \cite{eklund1992Neural,yager1994Analysis}, expert system \cite{ibanez2012apparel} and so on. A lot of research has been done on combining OWA operator with decision making. Yager has presented a measure based dominance which can order the uncertain criteria satisfactions and thus utilizing OWA operator to make the best decision \cite{yager2017owa}. Here, we further promote this method to fuzzy system, where the linguistic satisfactions are all fuzzy measures.

There are decision situations where the information cannot be evaluated precisely in a quantitative measure but may be in a qualitative one. For example, we may assess some criteria of the alternatives in linguistic terms such as "perfect" or "bad". It is inappropriate to set "perfect" as 1 and "bad" as 0 precisely. As a result, we transform the linguistic terms to fuzzy numbers due to the flexible framework which allows us to represent the information in a more adequate way. Since linguistic fuzzy satisfactions can be integrated into the expert system, and the generalization ability is not largely affected by the data, our proposed method is more universal and more consistent with the real world.

\subsection{Literature review}
In this subsection, we review some related work about decision making.

Many approaches has been applied to make decision for alternative suppliers' evaluation and selection \cite{Ho2010Multi}. Data envelopment analysis is often used in industry to measure the efficiencies the performance of alternative suppliers \cite{baker1997closer,liu2000using,b2001analytical,narasimhan2001supplier}, sometimes from both quantitative and qualitative aspects. Mathematical programming is also formulated to evaluate criteria, such as linear programming \cite{talluri2005note,talluri2003vendor,ng2008efficient}, goal programming \cite{karpak2001purchasing} and multi-objective programming \cite{narasimhan2006multiproduct,wadhwa2007vendor}. Analytic hierarchy process is a popular method to deal with the selection problem \cite{akarte2001web,chan2003interactive,muralidharan2002multi,chan2004development,ChenApplSci84563}. Besides, there are a few applications building the model based on fuzzy set theory to deal with the problem \cite{chen2006fuzzy,sarkar2006evaluation,florez2007strategic,Jiang2017FMEA,Xiao2018AHybridFuzzy}. Lots of other mechanisms which have the similar function will not be listed here \cite{chen2007bi,ramanathan2007supplier,saen2007new,sevkli2007application,ha2008hybrid,yang2006supplier,mendoza2008effective,kahraman2003multi,chan2007global,deng2018dependence,Xiao2018Multisensor,Xiao2018Similarity,sym10020046,CHATTERJEE2018101}.
All of these methods can be integrated to help decision maker select the best supplier.

Typically, for multi-criteria decision making, many novel thoughts are emerged, which are either evaluating the performance of the alternatives with respect to the criteria or finding the importance (weight) of the criteria. Ye and Jun took into account the unknown degree of interval-valued intuitionistic fuzzy sets to overcome the difficult decision of existing accuracy functions to the alternatives \cite{ye2009multicriteria}. Triantaphyllou and Sanchez tried to perform sensitivity analysis on the weights of the decision criteria \cite{triantaphyllou1997sensitivity}. Liu and Wang defined an evaluation function to measure the satisfactory degree to decision maker in intuitionistic fuzzy environment \cite{liu2007multi}. Yu et al. proposed two hesitant fuzzy linguistic harmonic averaging operators to solve the fuzzy linguistic decision making problem \cite{yu2018hesitant}. Tan and Chen applied fuzzy Choquet integral to aggregate criteria \cite{tan2011induced}, which is also discussed by Yager \cite{yager2018multi}. Yager also put forward possibilistic exceedance function and measure-based dominance to give a dominance relationship in uncertain decision making \cite{yager2017bidirectional}. Bordogna et al. used OWA operator to consider multi-person decision problem in a linguistic context \cite{bordogna1997linguistic}. Jiang and Wei proposed an intuitionistic fuzzy evidential power average aggregation operator, which both considers the uncertainty and aggregates the original intuitionistic fuzzy numbers to make a more reasonable decision \cite{jiang2018IJSS}.

\subsection{Organisation}
The paper is organized as follows. We first review the concept and properties of OWA operator in Section \ref{owa}. How to make a proper order for fuzzy number has been discussed in Section \ref{orderfn}. Section \ref{dominance} presents the dominance ordering method in multi-criteria decision making with uncertain satisfaction for OWA aggregation. An example is illustrated in section \ref{Example}. Finally, this paper is concluded in section \ref{Conclusion}.

\section{OWA aggregation}
\label{owa}
Ordered Weighted Averaging (OWA) operator, first introduced by Yager in 1988, is focus on aggregating multi-criteria to form an overall decision function.

\noindent \textbf{Definition: OWA operator.} Assume a mapping $OWA$ from $R^{n}\rightarrow R$, we say that $OWA$ is an OWA operator of dimension $n$ if it satisfies the following properties.

(1) $OWA(a_{1},a_{2},\ldots,a_{n})=\sum_{j=1}^{n}w_{j}b_{j}$, where $w=(w_{1},w_{2},\ldots,w_{n})^{T}$ is a correlated weighed vector.

(2) $w_{j}\in[0,1]$.

(3) $\sum_{j=1}^{n}w_{j}=1$.

(4) $b_{j}$ is the $j$th largest elements in $(a_{1},a_{2},\ldots,a_{n})$.

We can easily see that OWA operator has the following characteristics.

\noindent \textbf{Theorem 1: Monotonicity.} Suppose there are any two of vectors $(a_{1},a_{2},\ldots,a_{n})$ and $(b_{1},b_{2},\ldots,b_{n})$, if $a_{i}\leq b_{i}$ for all $i$, then $OWA(a_{1},a_{2},\ldots,a_{n})\leq OWA(b_{1},b_{2},\ldots,b_{n})$.

\noindent \textbf{Theorem 2: Commutativity.} If $(b_{1},b_{2},\ldots,b_{n})$ is an any substitute of $(a_{1},a_{2},\ldots,a_{n})$, then $OWA(a_{1},a_{2},\ldots,a_{n})=OWA(b_{1},b_{2},\ldots,b_{n})$.

\noindent \textbf{Theorem 3: Idempotency.} Suppose there is any vector $(a_{1},a_{2},\ldots,a_{n})$, we can get $OWA(a_{1},a_{2},\ldots,a_[n])=a$ if $a_{i}=a$ for all $i$. Particularly, if $W=W^{\ast}=(1,0,\ldots,0)^{T}$, then $OWA(a_{1},a_{2},\ldots,a_{n})=max_{i}(a_{i})$. If $W=W_{\ast}=(0,0,\ldots,1)^{T}$, then $OWA(a_{1},a_{2},\ldots,a_{n})=min_{i}(a_{i})$. If $W=W_{ave}=(\frac{1}{n},\frac{1}{n},\ldots,\frac{1}{n})^{T}$, then $OWA(a_{1},a_{2},\ldots,a_{n})=\frac{1}{n}\sum_{j=1}^{n}a_{j}$.

\noindent \textbf{Theorem 4: Extremum.} $min_{i}(a_{i})\leq OWA(a_{1},a_{2},\ldots,a_{n})\leq max_{i}(a_{i})$.

One of the advantages of OWA operator is that it provide a parameterized procedure for combining weight with criteria. All we need to consider is a rational ordering of the arguments. There are mainly three kinds of ways to calculate the weights, which are decision maker's non-subjective preference algorithm, decision maker's subjective preference algorithm and fuzzy semantic quantization algorithm.

Here we mainly focus on the third algorithm \cite{yager1996quantifier}. Drawing upon Zadeh's concept of linguistic quantifiers \cite{zadeh1975concept}, Yager came up with monotonic type quantifier $Q$ as function :$[0,1]\rightarrow[0,1]$ where $Q(0)=0$, $Q(1)=1$, and $Q(a)\leq Q(b)$ if all $a,b\in R$ where $a\leq b$ \cite{yager1996quantifier}. Furthermore he presented that the weights can be gained from $w_{j}=Q(\frac{j}{q})-Q(\frac{j-1}{q})$ for $j=1-q$. Function $Q$ can be related to linguistic terms, which is crucially important when it comes to fuzzy system. In this case all of the arguments have equal importance. If every arguments $a_{i}$ have different importance weights $\lambda_{i}$, then we can get the weights for $j=1-q$ as $w_{j}=Q(S_{j})-Q(S_{j-1})$ where $S_{j}=\sum_{k=1}^{j}\lambda_{\rho(k)}$ \cite{yager1997inclusion}.

\section{Ordering of triangular fuzzy number}
\label{orderfn}
Fuzzy number, first introduced by Zadeh in 1965, has applied in many domains to solve the uncertain circumstances problem \cite{zadeh1974fuzzy}. There are many kinds of fuzzy number, such as triangular fuzzy number, trapezoidal fuzzy number and other generalized fuzzy number.

\noindent \textbf{Definition: Generalized fuzzy number} In general, let $\widetilde{A}=(a,b,c,d;\omega)$ be a fuzzy number, whose membership function can be defined as follows.

$$
f_{\widetilde{A}}(x)=
\left\{
\begin{array}{rcl}
  f_{\widetilde{A}}^{L}(x)& &{a\leq x\leq b}\\
  \omega& & {b\leq x\leq c}\\
  f_{\widetilde{A}}^{R}(x)& & {c\leq x\leq d}\\
  0& & {otherwise}
\end{array}\right .$$
where $0<\omega \leq 1$ is a constant, $f_{\widetilde{A}}^{L}(x):[a,b]\rightarrow[0,\omega]$ and $f_{\widetilde{A}}^{R}(x):[c,d]\rightarrow[0,\omega]$ are two strictly monotonic and continuous mappings from R to the closed interval $[0,\omega]$.

\begin{figure}[!ht]
  \centering
  \includegraphics[height=6cm]{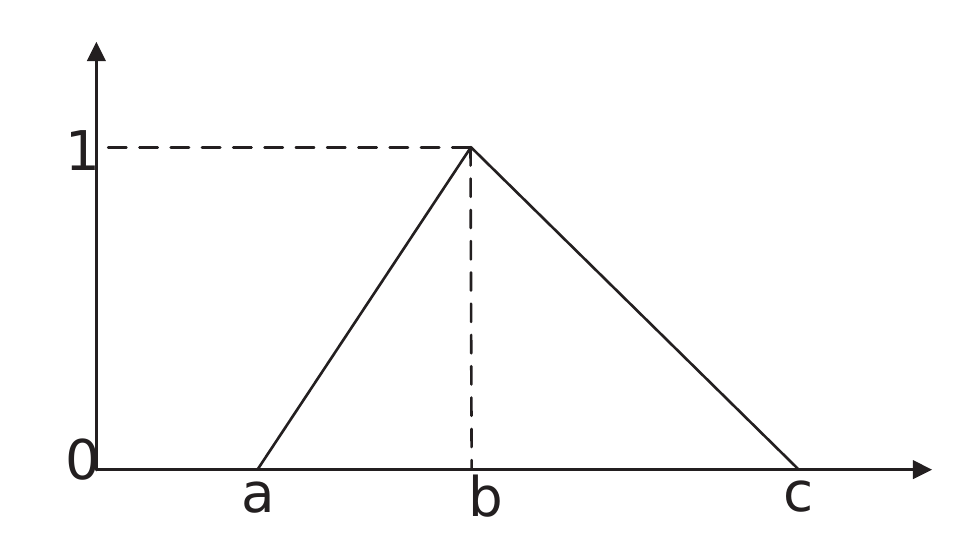}
  \caption{Triangular fuzzy number according to the membership function}
     \label{trifuzzynum}
\end{figure}

Here we mainly focus on triangular fuzzy number, which is shown in figure \ref{trifuzzynum}. We call $\widetilde{A}=(a,b,c)$ a triangular fuzzy number whose membership function is

$$
f_{\widetilde{A}}(x)=
\left\{
\begin{array}{rcl}
  0& &{x\leq a}\\
  \frac{x-a}{b-a}& & {a<x\leq b}\\
  \frac{x-c}{b-c}& & {b<x\leq c}\\
  0& & {x>c}
\end{array}\right .$$

Since the real sets are linearly ordered, which is not the case for fuzzy set, how to compare two fuzzy number has always been debated. There are several approaches to give an order of fuzzy number, which generally fall under two main categories of either building a fuzzy-real sets mapping or defining a dominance relation of one fuzzy set over another. In the following subsections, we discuss 2 popular methods  for ranking fuzzy numbers from different categories separately.


\subsection{A fuzzy-real sets mapping: Centroid-based distance method}
Here we only consider positive fuzzy numbers. Cheng suggested a centroid-based distance method for ranking fuzzy number, which utilizes the Euclidean distances from the origin to the centroid point of each fuzzy number to compare and rank the fuzzy numbers.

In \cite{wang2006centroids}, Wang et al. proposed that the correct centroid formulae should be as follows:
\begin{eqnarray}
\label{xfnum}
\overline{x}_{0}(\widetilde{A})=\frac{\int_{-\infty}^{+\infty}xf_{\widetilde{A}}(x)dx}{\int_{-\infty}^{+\infty}f_{\widetilde{A}}(x)dx}=\frac{\int_{a}^{b}xf_{\widetilde{A}}^{L}(x)dx+\int_{b}^{c}(x\omega)dx+\int_{c}^{d}xf_{\widetilde{A}}^{R}(x)dx}{\int_{a}^{b}f_{\widetilde{A}}(x)dx+\int_{b}^{c}(\omega)dx+\int_{c}^{d}f_{\widetilde{A}}^{R}(x)dx}
\end{eqnarray}
\begin{eqnarray}
\label{yfnum}
\overline{y}_{0}(\widetilde{A})=\frac{\int_{0}^{\omega}y(g_{\widetilde{A}}^{R}(y)-g_{\widetilde{A}}^{L}(y))dy}{\int_{0}^{\omega}(g_{\widetilde{A}}^{R}(y)-g_{\widetilde{A}}^{L}(y))dy}
\end{eqnarray}
where $g_{\widetilde(A)}^{L}:[0,\omega]\rightarrow[a,b]$ and $g_{\widetilde(A)}^{R}:[0,\omega]\rightarrow[c,b]$ are the inverse functions of $f_{\widetilde{A}}^{L}(x)$ and $f_{\widetilde{A}}^{R}(x)$, respectively.

Since triangular fuzzy numbers are special cases of fuzzy numbers, for any triangular fuzzy number $\widetilde{A}=(a,b,c)$ with a piecewise linear membership function, its centroid can be determined by
\begin{eqnarray}
\label{xtfnum}
\overline{x}_{0}(\widetilde{A})=\frac{1}{3}(a+b+c)
\end{eqnarray}
\begin{eqnarray}
\label{ytfnum}
\overline{y}_{0}(\widetilde{A})=\frac{1}{3}
\end{eqnarray}

Therefore, ranking fuzzy numbers can be transformed into ordering the centroid of fuzzy numbers.
\subsection{A dominance relation of one fuzzy set over another: Lattice operators and inclusion index}
\noindent \textbf{Definition: The fuzzy lattices.}
In \cite{klir1996fuzzy}, Klir and Yuan propose that we can order the fuzzy numbers by extending the real numbers lattice operations $min$ and $max$ to corresponding fuzzy lattice operation $MIN$ and $MAX$, which is defined as follows.
\begin{eqnarray}
\label{minmax1}
MIN(A,B)(z)=\underset{z=min(x,y)}{sup} min[A(x),B(y)]\
\end{eqnarray}
\begin{eqnarray}
\label{minmax2}
MAX(A,B)(z)=\underset{z=max(x,y)}{sup} min[A(x),B(y)]
\end{eqnarray}
where $x,y,z\in R$.

\noindent \textbf{Theorem 1 (Chiu-Wang-2002).} For any triangular fuzzy numbers A and B, defined on the universal set R, with continuous membership function and $(A\cap B)(x_{m})\geq(A\cap B)(x)$ for all $x\in R$ and $A(x_{m})=B(x_{m})$, moreover, $x_{m}$ is betwen two mean values of $A$ and $B$ (if the number of $x_{m}$ is not unique, any one point of those $x_{m}$ is suitable). Then the operation $MIN$ can be implemented as

$$
MIN(A,B)(z)=
\left\{
\begin{array}{rcl}
  (A\cup B)(z)& &{as\ z<x_{m}}\\
  (A\cap B)(z)& &{as\ z\geq x_{m}}
\end{array}\right .$$
where $x\in Z=R$, and $\cup$, $\cap$ denote the standard fuzzy intersection and union, respectively.

\noindent \textbf{Definition: Inclusion index (InI).} For a discrete and finite set A and its membership function $\mu_{A}\in[0,1]$, the absolute cardinality can be formulated as $Card(A)=|A|=\sum_{x\in X}^{}\mu_{A}(x)$ with $A\subseteq X$. The inclusion index of discrete sets is
\begin{eqnarray}
\partial(E\subseteq F)=\frac{\sum\parallel E\cap F\parallel}{\sum\parallel E\parallel}=\frac{\sum_{x\in X}^{}T(\mu_{E}(x),\mu_{F}(x))}{\sum_{x\in X}^{}\mu_{E}(x)}
\end{eqnarray}
Similarly, for a continuous and finite set B and its membership function $\mu_{B}\in[0,1]$, the absolute cardinality can be formulated as $Card(B)=|B|=\int_{x\in X}^{}\mu_{B}(x)dx$ with $B\subseteq X$. The inclusion index of continuous sets is
\begin{eqnarray}
\partial(E\subseteq F)=\frac{\int\parallel E\cap F\parallel}{\int\parallel E\parallel}=\frac{\int_{x\in X}^{}T(\mu_{E}(x),\mu_{F}(x))}{\int_{x\in X}^{}\mu_{E}(x)}
\end{eqnarray}
where T is a triangular norm and $\parallel$ denote the standard fuzzy cardinal operator $Card$, $\mu_{E}$ and $\mu_{F}$ are the membership function of $E$ and $F$. respectively.

In \cite{boulmakoul2013ranking}, Boulmakoul et al. introduced a novel ranking operator "$\prec$", "$\succ$" and "$\simeq$" for every fuzzy sets A and B by the following implications:

If MIN $\in \{A,B\}$
\begin{eqnarray}
\left\{\begin{array}{cc}
  A\prec B \Leftrightarrow MIN=A\\
  A\succ B \Leftrightarrow MIN=B\\
  A\simeq B \Leftrightarrow MIN=A\ and\ MIN=B\\
\end{array}\right .
\end{eqnarray}

Else
\begin{eqnarray}
\left\{\begin{array}{cc}
  A\prec B \Leftrightarrow \partial(MIN\subseteq A)>\partial(MIN\subseteq B)\\
  A\succ B \Leftrightarrow \partial(MIN\subseteq A)<\partial(MIN\subseteq B)\\
  A\simeq B \Leftrightarrow \partial(MIN\subseteq A)=\partial(MIN\subseteq B)\\
\end{array}\right .
\end{eqnarray}

Such operator can provide a reasonable order of fuzzy numbers, which is a broad level of dominance relation.
\section{Dominance ordering in multi-criteria decision making with uncertain satisfaction}
\label{dominance}
Assume that there is a collection of criteria $C=\{C_{1},C_{2},\ldots,C_{q}\}$ which need to be considered for decision maker. For a specific alternative $x$, using OWA operator we can calculate the whole satisfaction that $D(x)=OWA(C_{1}(x),C_{2}(x),\ldots,C_{q}(x))=\sum_{i=1}^{q}w_{i}C_{\rho(i)}$, where $C_{k}(x)$ is the satisfaction degree for criteria $C_{k}$ and $C_{\rho(i)}$ is the $i$th largest in $C_{k}(x)$.

In the following, we assume that $Y=\{y_{1},y_{2},\ldots,y_{n}\}$ is the set of the criteria satisfaction values to decision maker. Each $y_{i}$ is a fuzzy number $(a,b,c)$ with $0\leq a\leq b\leq c\leq1$ and $y_{j}>y_{j-1}$. Since criteria has some uncertainty, and there are numerous methods to assess the uncertainty information, in \cite{yager2017owa} Yager took into account three common cases: probability distribution, possibility distribution and interval values. Different cases has different types of uncertainty that need specific formulations to model the uncertainty information.

(1) The uncertain value $C_{k}(x)$ is a probability distribution $P_{k}=[p_{k1},\ldots,p_{kj},\ldots,p_{kn}]$ on Y, where $p_{kj}$ is the probability distribution of satisfaction of $C_{k}$ by $x$ is fuzzy number $y_{j}$, $p_{k,j}\in [0,1]$ and $\sum_{j=1}^{n}p_{kj}=1$.

(2) The uncertain value $C_{k}(x)$ is a possibility distribution $\Pi_{k}=[\tau_{k1},\ldots,\tau_{kj},\ldots,\tau_{kn}]$ on Y, where $\tau_{kj}$ is the possibility distribution of satisfaction of $C_{k}$ by $x$ is fuzzy number $y_{j}$, $\tau_{kj}\in [0,1]$ and $Max_{j}[\tau_{kj}]=1$.

(3) The uncertain value $C_{k}(x)$ is an interval value $C_{k}(x)=[a_{k},b_{k}]$, then the possibility distribution is that $\tau_{kj}=1$ for each $y_{j}\in [a_{k},b_{k}]$ and $\tau_{kj}=0$ for all $y_{j}\notin [a_{k},b_{k}]$.

Then monotonic set measures are implemented to provided a unified framework for corresponding the uncertain value to our knowledge of $C_{k}(x)$.

\noindent \textbf{Definition: Monotonic set measures (fuzzy measures).} A monotonic set measure $\mu$ is a mapping: $\mu:2^{Y}\rightarrow[0,1]$ with $\mu(\emptyset)=0$, $\mu(Y)=1$ and $\mu(A)\geq \mu(B)$ for $A\supseteq B$. For any subset $A$ of $Y$, we denote $\mu_{k}(A)=C_{k}(A)$ as the anticipation of finding $C_{k}(x)$ in $A$. It is clear that the cardinality of subset A has a positive correlation with $\mu_{k}(A)$ \cite{yager2017owa}.

(1) If $C_{k}(x)$ is a probability distribution $P_{k}=[p_{k1},\ldots,p_{kj},\ldots,p_{kn}]$ on Y, then $\mu_{k}({y_{j}})=p_{kj}$ and $\mu_{k}(A)=\sum_{y_{j}\in A}^{}\mu_{k}({y_{j}})=\sum_{y_{j}\in A}^{}p_{kj}$.

(2) If $C_{k}(x)$ is a possibility distribution $\Pi_{k}=[\tau_{k1},\ldots,\tau_{kj},\ldots,\tau_{kn}]$ on Y, then $\mu_{k}({y_{j}})=\tau_{kj}$ and $\mu_{k}(A)=\underset{y_{j}\in A}{Max}[\mu_{k}({y_{j}})]=\underset{y_{j}\in A}{Max}[\tau_{kj}]$.

(3) If $C_{k}(x)$ is an interval value $C_{k}(x)=[a_{k},b_{k}]$, then $\mu_{k}(y_{j})=1$ for $y_{j}\in [a_{k},b_{k}]$, $\mu_{k}(y_{j})=0$ for $y_{j}\notin [a_{k},b_{k}]$ and $\mu_{k}(A)=\underset{y_{j}\in A}{Max}[\mu({y_{j}})]$.

For any alternative $x$, in order to rank $C_{\rho(i)}$, Yager introduced the concept of measure-based dominance\cite{yager2014stochastic}.

\noindent \textbf{Definition: Measure-based dominance.} If $H_{j}=\{y_{1},\ldots,y_{j}\}$ is the subset of $j$ largest elements in $Y$, then $C_{k_{1}}(x)$ dominates $C_{k_{2}}(x)$ if $\mu_{k_{1}}(H_{j})\geq\mu_{k_{2}}(H_{j})$ for all $j$ and $\mu_{k_{1}}(H_{j})>\mu_{k_{2}}(H_{j})$ for at least one $H_{j}$. We denote that $C_{k_{1}}(x)>_{\mu D} C_{k_{2}}(x)$, meaning that $C_{k_{1}}(x)$ seems bigger than $C_{k_{2}}(x)$.

Since $>_{\mu D}$ is a pairwise relationship, it has the properties of transitivity and completeness. As a result, the ordering of $C_{k}(x)$ can be obtained based on the relationship $>_{\mu D}$. Thus we can calculate the OWA aggregation of the $C_{k}(x)$ that

\begin{eqnarray}
OWA(C_{1},\ldots,C_{k}(x),\ldots,C_{q}(x))=\sum_{i=1}^{q}w_{i}C_{\rho(i)}(x)=\sum_{i=1}^{q}w_{i}\mu_{\rho(i)}
\end{eqnarray}

In this case, $\mu$ is defined as a measure that aggregation $\mu_{\rho(j)}$ for all j such that $\mu=OWA(C_{1},\ldots,C_{k}(x),\ldots,C_{q}(x))$. For all $H_{j}=\{y_{i}|for\ i=1-j\}$ where $H_{j}\subseteq Y$ we have $\mu(H_{j})=\sum_{j=1}^{q}w_{j}\mu_{\rho(j)}(A)$.

However, there are situations such as $\mu_{1}$ is the biggest in $\mu(H_{1})$ and $\mu_{2}$ is the biggest in $\mu(H_{2})$ and $\mu_{3}$ is the biggest in $\mu(H_{3})$, thus we cannot find the complete dominance relationship for all distributions. To overcome this difficulty, in \cite{yager2014stochastic} Yager proposed a surrogate for dominance ordering based on the Choquet integral. $M(\mu_{k})$ is defined as a surrogate for $\mu_{k}$ such that $M(\mu_{k})=\sum_{j=1}^{n}(\mu_{k}(H_{j})-\mu_{k}(H_{j-1}))y_{j}$, where $\mu_{k}(H_{j})-\mu_{k}(H_{j-1})$ can be seen as a set of weights of the $y_{j}$. $M(\mu_{k})$ contains two desirable properties:

\noindent \textbf{(1) Certainty.} $M(\mu_{k})=y_{k}$ if $\mu_{k}$ is a certian measure focused at $y_{k}$.

\noindent \textbf{(2) Consistency with dominance.} $M(\mu_{1})>M(\mu_{2})$ if $\mu_{1}>_{\mu D}\mu_{2}$.

Since $\mu_{k}(H_{j})-\mu_{k}(H_{j-1})$ is a certain measure, $M(\mu_{k})$ is a fuzzy number due to $y_{i}$. We have been discussed how to give a reasonable ordering of the fuzzy number in section \ref{orderfn}, therefore we can get the dominance ordering of $C_{\rho(i)}$.

\begin{figure}[!ht]
  \centering
  \includegraphics[height=16cm]{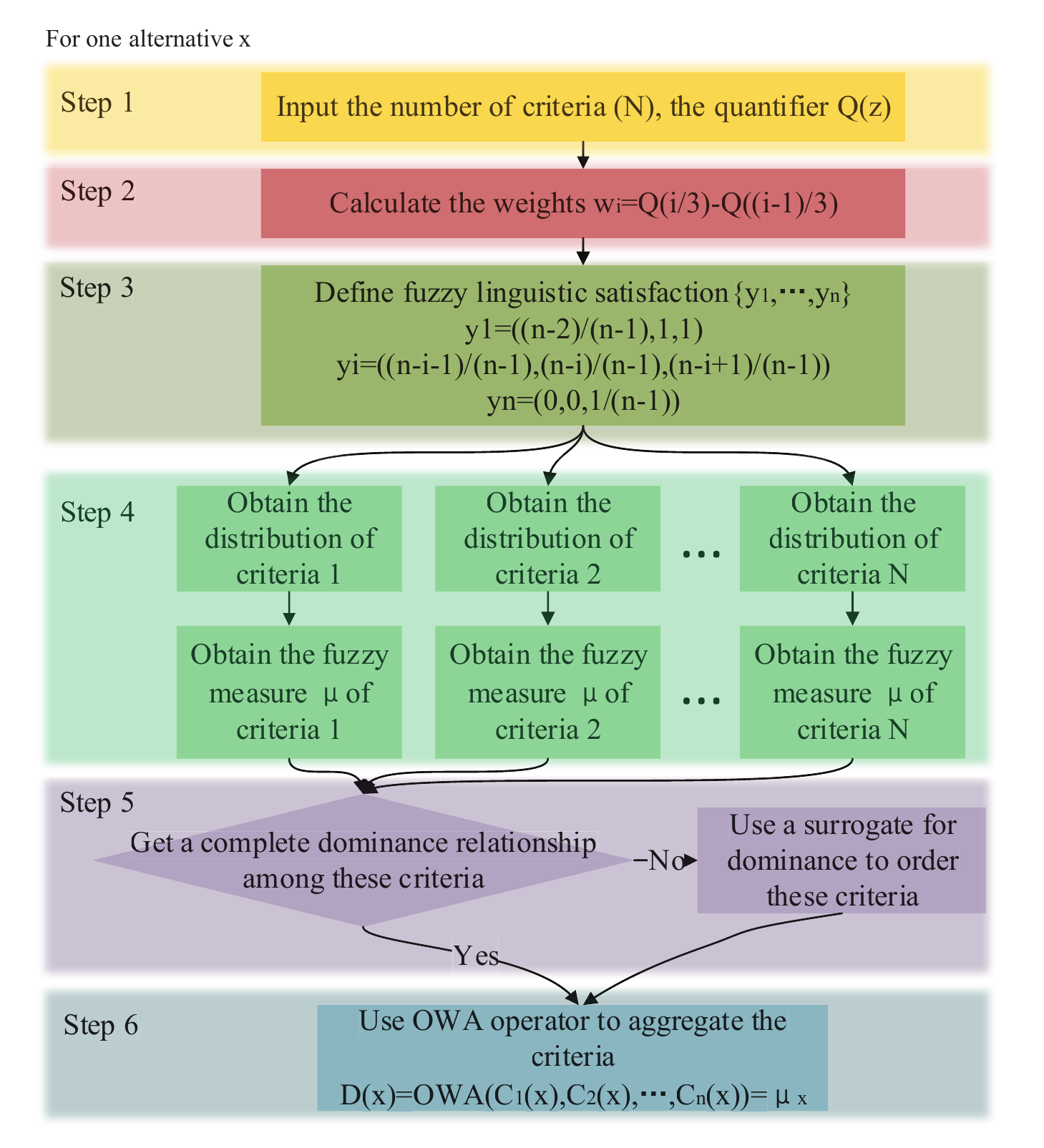}
  \caption{The whole procedure of multi-criteria decision making with uncertain fuzzy satisfaction}
     \label{procedure}
\end{figure}

For each alternative, we can get an overall score from the whole procedure of multi-criteria decision making with uncertain fuzzy satisfaction, which is shown in figure \ref{procedure}. Our final option is to choose the best alternative based on the scores.

\section{Example}
\label{Example}
An example is shown to illustrate our approach. Assume there are three criteria $C=\{C_{1}, C_{2}, C_{3}\}$. Assume that there exists some uncertainty about our knowledge of the $C_{k}(x)$'s. To be more specific, let's suppose that $C_{1}$ is a probability distribution P where

$p_{1}=0.5, p_{2}=0.3, p_{3}=0.2, p_{4}=0, p_{5}=0$

\noindent $C_{2}$ is a possibility distribution T where

$\tau_{1}=0.2, \tau_{2}=1, \tau_{3}=0.8, \tau_{4}=0.1, \tau_{5}=0$

\noindent $C_{3}$ is a precise knowledge where $C_{3}(x)=y_{4}=0.6$

Since our target is to satisfy "most" of the criteria, for alternative $x$, we have

\begin{eqnarray}
D(x)=OWA(C_{1}(x),C_{2}(x),C_{3}(x))=\sum_{i=1}^{3}w_{i}C_{\rho(i)}
\end{eqnarray}

where $w_{i}=Q(\frac{i}{3})-Q(\frac{i-1}{3})$ for $i=1,2,3$. $Q(z)=z^{2}$ is defined as quantifier Q to represent the degree of "most".

Then, we can get $w_{1}=(\frac{1}{3})^{2}-0=0.11, w_{1}=(\frac{2}{3})^{2}-(\frac{1}{3})^{2}=0.33, w_{1}=1-(\frac{2}{3})^{2}=0.56$

Let $Y=\{y_{1},y_{2},y_{3},y_{4},y_{5}\}$. Here we correspond linguistic satisfactions \{perfect, large, moderate, small, none\} to triangular fuzzy numbers. To be more specific, we define the correlation as follows.

Perfect: $y_{1}=\{0.75,1,1\}$.

Large: $y_{2}=\{0.5,0.75,1\}$.

Moderate: $y_{3}=\{0.25,0.5,0.75\}$.

Small: $y_{4}=\{0,0.25,0.5\}$.

None: $y_{5}=\{0,0,0.25\}$.

For $C_{1}(x)$, we have a probability measure $\mu_{1}$:

\noindent $\mu_{1}(\{y_{1}\})=0$, $\mu_{1}(\{y_{2}\})=0.2$, $\mu_{1}(\{y_{3}\})=0.5$, $\mu_{1}(\{y_{4}\})=0.2$, $\mu_{1}(\{y_{5}\})=0.1$.

For $C_{2}(x)$, we have a possibility measure $\mu_{2}$:

\noindent $\mu_{2}(\{y_{1}\})=0.4$, $\mu_{2}(\{y_{2}\})=0.2$, $\mu_{2}(\{y_{3}\})=0.6$, $\mu_{2}(\{y_{4}\})=0.8$, $\mu_{1}(\{y_{5}\})=1$.

For $C_{3}(x)$, we have a certain measure $\mu_{3}$ with $\mu_{3}(A)=1$ if $y_{3}\in A$ and $\mu_{3}(A)=0$ if $y_{3}\notin A$.

Define $H_{j}=\{y_{1},y_{2},\ldots,y_{j}\}$ and we can obtain Table \ref{muk}.

\begin{table}[htbp]
\small
  \centering
  \caption{Values of $\mu_k(H_{j})$}\label{muk}
    \begin{tabular}{cccccc}
    \toprule[1.5pt]
      \ &$\mu_{k}(H_{1})$ & $\mu_{k}(H_{2})$ & $\mu_{k}(H_{3})$ & $\mu_{k}(H_{4})$ & $\mu_{k}(H_{5})$  \\
      \hline
      $\mu_{1}$& 0.0 & 0.2 & 0.7 & 0.9 & 1.0  \\
      $\mu_{2}$& 0.4 & 0.4 & 0.6 & 0.8 & 1.0 \\
      $\mu_{3}$& 0.0 & 0.0 & 0.0 & 1.0 & 1.0 \\
    \bottomrule[1.5pt]
    \end{tabular}%
\end{table}%

The first step is to decide whether an ordering of $\mu_{k}(H_{j})$ can satisfy dominance, meaning that for each $j$ the ordering of $\mu_{k}(H_{j})$ has to be the same. It turns out that our situation does not fit dominance, because $\mu_{2}$ is bigger for j=1, $\mu_{1}$ is bigger for j=3 and $\mu_{3}$ is bigger for j=4. To order the $C_{k}(x)$, we need to use surrogate formulas $M(\mu_{k})=\sum_{j=1}^{5}(\mu_{k}(H_{j})-\mu_{k}(H_{j-1}))y_{j}$. Denoting that $\mu_{k}(H_{0})=0$, we calculate $V_{kj}=\mu_{k}(H_{j})-\mu_{k}(H_{j-1})$ and the result is shown in Table \ref{vkj}.

\begin{table}[htbp]
\small
  \centering
  \caption{$V_{kj}=\mu_{k}(H_{j})-\mu_{k}(H_{j-1})$}\label{vkj}
    \begin{tabular}{cccccc}
    \toprule[1.5pt]
      \ &$V_{k1}$ & $V_{k2}$ & $V_{k3}$ & $V_{k4}$ & $V_{k5}$  \\
      \hline
      $\mu_{1}$& 0.0 & 0.2 & 0.5 & 0.2 & 0.1  \\
      $\mu_{2}$& 0.4 & 0.0 & 0.2 & 0.2 & 0.2 \\
      $\mu_{3}$& 0.0 & 0.0 & 0.0 & 1.0 & 0.0 \\
    \bottomrule[1.5pt]
    \end{tabular}%
\end{table}%

Here then $M_(\mu_{k})=\sum_{j=1}^{5}V_{kj}y_{j}=V_{k1}\times (0.75,1,1)+V_{k2}\times (0.5,0.75,1)+V_{k3}\times (0.25,0.5,0.75)+V_{k4}\times (0,0.25,0.5)+V_{k5}\times (0,0,0.25)$. In this case, $M(\mu_{1})=(0.225,0.45,0.7)$, $M(\mu_{2})=(0.35,0.55,0.7)$, $M(\mu_{3})=(0,0.25,0.5)$.

There are lots of methods to order the fuzzy number. Here we calculate the fuzzy centroid to order the fuzzy numbers. The fuzzy centroid is defined as $\overline{x}_{0}(\widetilde{A})=\frac{a+b+c}{3}$, where the fuzzy number is $\widetilde{A}=(a,b,c)$. Thus, the fuzzy centroid for k=1-3 is that $\overline{x}_{0}(\mu_{1}) =0.458$, $\overline{x}_{0}(\mu_{2}) =0.533$, $\overline{x}_{0}(\mu_{3}) =0.25$. Since $\overline{x}_{0}(\mu_{2})>\overline{x}_{0}(\mu_{1})>\overline{x}_{0}(\mu_{3})$, our ordering is $M(\mu_{2})>M(\mu_{1})>M(\mu_{3})$ and thus we obtain

$D(x)=w_{1}C_{2}(x)+w_{2}C_{1}(x)+w_{3}C_{3}(x)$

We define $\mu_{x}$ as a measure of each $D(x)$ on $Y$, where $\mu_{A}=w_{1}\mu_{2}(A)+w_{2}\mu_{1}(A)+w_{3}\mu_{3}(A)$ for all $A\subseteq Y$. According to the previously gained OWA weighing factors $w_{i}$, we obtain

$\mu_{x}(A)=0.11\mu_{2}(A)+0.33\mu_{1}(A)+0.56\mu_{3}(A)$

It is the same as $H_{i}$. For $H_{i}\subseteq Y$, we obtain
\begin{eqnarray}
\label{muhi}
\mu_{x}(H_{i})=0.11\mu_{2}(H_{i})+0.33\mu_{1}(H_{i})+0.56\mu_{3}(H_{i})
\end{eqnarray}

Therefore, for j=1-5, we can get $\mu_{x}(H_{j})$ based on Table \ref{muk} and formula \ref{muhi}.

\noindent $\mu_{x}(H_{1})=0.044$, $\mu_{x}(H_{2})=0.066$, $\mu_{x}(H_{3})=0.187$, $\mu_{x}(H_{4})=0.648$, $\mu_{x}(H_{5})=0.055$

Here the $D(x)=\mu_{x}$ is an OWA aggregation where $\mu_{x}$ is a measure on $Y$, representing the aggregation of multi-criteria with fuzzy satisfaction of alternative $x$. In a real world, we will have to select the most satisfactory choice from a collection of possible alternatives $X=\{x_{1},\ldots,x_{i},\ldots,x_{N}\}$. Using the previous procedure we can gain $D(x_{i})=\mu_{x_{i}}$ for each $\mu_{x_{i}}$. Then our final choice $x^{\ast}$ has the largest $D(x^{\ast})$. We still need to use dominance relationship to order $\mu_{x_{i}}$, trying to find whether there is one $\mu_{x_{i}}$ dominating all the others. If so, such $x_{i}$ is our final alternative. If not, we need to use the surrogate method and calculate $M(\mu_{x_{i}})=\sum_{j=1}^{n}(\mu_{x_{i}}(H_{j})-\mu_{x_{i}}(H_{j-1}))y_{j}$ for each $\mu_{x_{i}}$. Then we can order $x_{i}$ using $M(\mu_{x_{i}})$ and thus choosing the best alternative.

\section{Conclusion}
\label{Conclusion}
This paper promotes the OWA aggregation of multi-criteria with mixed uncertain satisfactions in linguistic fuzzy measures. We first review the Ordered Weighed Averaging (OWA) operator and the fuzzy number. Then we discuss the mechanisms to rank the fuzzy numbers rationally. To aggregate the multi-criteria, we utilize measure-based dominance and a surrogate for dominance to gain the order of the uncertain arguments. Finally, we provide the whole procedure of the approach and present an example to illustrate it.

\section*{Acknowledgment}
 The work is partially supported by National Natural Science Foundation of China (Grant Nos. 61573290, 61503237).

\section{Reference}
\label{Reference}

\bibliographystyle{unsrt}
\bibliography{owaref}
\end{document}